\documentclass[sigconf, nonacm]{acmart}

\usepackage{amsmath}
\usepackage{algorithm}
\usepackage[noend]{algpseudocode}
\usepackage{xcolor}
\usepackage{svg}
\usepackage{parskip}
\usepackage{multicol}
\usepackage{subfigure}

\algrenewcommand\algorithmicrequire{\textbf{Input:}}
\algrenewcommand\algorithmicensure{\textbf{Output:}}

\newtheorem{theorem}{Theorem}[section]

\newcommand{\squishlist}{
 \begin{list}{$\bullet$}
  { \setlength{\itemsep}{0pt}
     \setlength{\parsep}{1pt}
     \setlength{\topsep}{1pt}
     \setlength{\partopsep}{0pt}
     \setlength{\leftmargin}{1.5em}
     \setlength{\labelwidth}{1em}
     \setlength{\labelsep}{0.5em} } }
\newcommand{\squishend}{\end{list}}

\AtBeginDocument{%
  \providecommand\BibTeX{{%
    \normalfont B\kern-0.5em{\scshape i\kern-0.25em b}\kern-0.8em\TeX}}}

\begin{document}

\title{OPTWIN: Drift identification with optimal sub-windows}

\author{Mauro Dalle Lucca Tosi}
\email{mauro.dalleluccatosi@uni.lu}
\orcid{0000-0002-0218-2413}
\affiliation{%
  \institution{University of Luxembourg}
  \city{Esch-sur-Alzette}
  \country{Luxembourg}
}

\author{Martin Theobald}
\email{martin.theobald@uni.lu}
\orcid{0000-0003-4067-7609}
\affiliation{%
  \institution{University of Luxembourg}
  \city{Esch-sur-Azette}
  \country{Luxembourg}}

\renewcommand{\shortauthors}{Tosi and Theobald}

\begin{abstract}
Online Learning (OL) is a field of research that is increasingly gaining attention both in academia and industry. One of the main challenges of OL is the inherent presence of {\em concept drifts}, which are commonly defined as unforeseeable changes in the statistical properties of an incoming data stream over time. The detection of concept drifts typically involves analyzing the {\em error rates} produced by an underlying OL algorithm in order to identify if a concept drift occurred or not, such that the OL algorithm can adapt accordingly. Current concept-drift detectors perform very well, i.e., with {\em low false negative rates}, but they still tend to exhibit {\em high false positive rates} in the concept-drift detection. This may impact the performance of the learner and result in an undue amount of computational resources spent on retraining a model that actually still performs within its expected range. In this paper, we propose OPTWIN, our ``OPTimal WINdow'' concept drift detector. OPTWIN uses a sliding window of events over an incoming data stream to track the errors of an OL algorithm. The novelty of OPTWIN is to consider {\em both the means and the variances} of the error rates produced by a learner in order to split the sliding window into two provably optimal sub-windows, such that the split occurs at the earliest event at which a statistically significant difference according to either the $t$- or the $f$-tests occurred. We assessed OPTWIN over the MOA framework, using ADWIN, DDM, EDDM, STEPD and ECDD as baselines over 7 synthetic and real-world datasets, and in the presence of both sudden and gradual concept drifts. In our experiments, we show that OPTWIN surpasses the F1-score of the baselines in a statistically significant manner while maintaining a lower detection delay and saving up to 21\% of time spent on retraining the models.
\end{abstract}



\received{01 February 2023}

\maketitle

\section{Introduction}
Online Learning (OL) is an area of research which gained an increasing amount of attention both in academia and industry over the past years. OL is a sub-field of Machine Learning (ML) in which an underlying ML technique aims to constantly update its model's parameters from an incoming data stream under limited time and space constraints. Ideally, OL methods are trained in real time and thereby aim to maximize the prediction accuracy of their models based on bounded windows of data instances they have previously seen, and which they may see in the near future \cite{hoi2021online}. Common use-cases of OL include social-networks analyses, various kinds of IoT-related services and predictions, spam detection, online fraud detection (e.g., credit-card transactions), and many others. 


The presence of {\em concept drifts} is one of the numerous challenges when processing data streams in real-time. Concept drifts are commonly defined as unforeseeable changes in the statistical properties of the incoming data stream over time \cite{lu2018learning}. Such a change may have different sources and types, and it may involve different adaptation strategies (cf. Section \ref{sec:background}). In practice, concept drift is the key phenomenon that impairs the performance of pre-trained (and hence static) ML models over data streams. Take as an example the spam-detection use-case studied by Fdez-Riverola et al. \cite{fdez2007applying}. Spammers are known to constantly adapt their spamming strategy (thus creating concept drifts intentionally) to overcome the spam filters. Therefore, if we train a spam filter based on a pre-defined dataset, it will perform as expected just until the spammers create a new form of spam that bypasses the pre-trained filter. Thus, if the pre-trained filter fails to detect a concept drift and, consequently, adapt itself accordingly, its performance may be drastically impacted.

To detect concept drifts, one may rely on one of the following two principal types of drift detectors (or on a combination of them): (1) {\em error rate-based}, which use the difference between the number or magnitude of historical prediction errors and new prediction errors to determine if a concept drift occurred \cite{lu2018learning}; and (2) {\em data distribution-based}, which use distance metrics to determine if the distribution from the historical data instances and the new data instances statistically changed \cite{lu2018learning}. 
We here highlight two fundamental differences between both approaches. First, as prediction errors already provide a form of aggregated measure for the performance of an ML model, it is usually computationally more expensive to track changes on features derived from the data stream (data distribution-based) rather than on the error rates (error rate-based). Second, error-based approaches need labels from the ML models, whereas data distribution-based ones do not. Note that, error-based approaches use labels from the ML problem to calculate the error rate of the ML models; thus, no concept drift-related labels are assumed to be available. This paper focuses on the drift identification of supervised ML problems, which already assume the availability of labels for training the ML models. Therefore, considering its lightweight performance compared to the data distribution-based approaches, we further focus on error rate-based methods. 

The most popular among the error rate-based algorithms are ADWIN \cite{bifet2007learning}, DDM \cite{gama2004learning}, and their variations as \cite{baena2006early, barros2017rddm}. ADWIN maintains a variable window $W$ of the past error rates produced by the learner. Then, it checks, at each iteration, if there are two sub-windows $W = W_0\,W_1$ whose absolute difference among their {\em mean values} $\mu_{W_0}$, $\mu_{W_1}$ of errors exceeds a given threshold $\epsilon$. If it finds two such sub-windows, with $|\mu_{W_0} - \mu_{W_1}| \geq \epsilon$, a drift is flagged and $W$ is shrunk to discard the staled data. DDM, on the other hand, tracks the error rate $p$ of the learner by its {\em standard deviation} $s$. A drift is flagged at iteration $i$ iff the current error rate $p_i$ surpasses the minimum previously recorded error rate $p_{min}$ by over 3 times its minimum standard deviation $s_{min}$, i.e., $p_i + s_i \geq p_{min} + 3\,s_{min}$. Both methods use the errors' standard deviations only to set their thresholds and guarantee the statistical significance of the errors' differences. Thus, they do not interpret changes in those standard deviations as potential concept drifts.

As opposed to previous approaches, we argue that also changes in the standard deviations of past prediction errors should trigger concept drifts. Take, as a simple example, a regressor that outputs the following errors at window $W_0$: 
$$W_0 = \langle 0.3;0.7;0.7;0.3;0.3;0.7;0.5;0.5 \rangle$$ 
While, at window $W_1$, it outputs the following errors:
$$W_1 = \langle 0.0;1.0;1.0;0.0;1.0;0.0;0.0;1.0\rangle$$
Current drift detectors like ADWIN would not consider this as a concept drift, as the error means $\mu_{W_0}$, $\mu_{W_1}$ over both windows are equal to 0.5, whereas we---intuitively---understand that something changed and, therefore, a concept drift should be flagged. 

In this paper, we propose the ``OPTimal WINdow'' drift detector (OPTWIN). It is also a sliding-window algorithm which analyzes the error rates (either binary or real-valued) produced by an underlying ML algorithm. It calculates the \textit{optimal cut} of a sliding window $W$ to divide it into two sub-windows $W_{\mathit{hist}}$ and $W_{\mathit{new}}$. 
It then performs the well-known $t$- (for means) and $f$-tests (for standard deviations) to determine whether the two sub-windows exhibit a statistically significant difference in either their means or standard deviations, respectively. OPTWIN has the following two main features: (1) it can calculate the \textit{optimal cut} based only on the length of $W$; (2) it uses both the errors' means and  standard deviations to flag drifts. Furthermore, we provide detailed and rigorous guarantees of OPTWIN's performance in terms of its true-positive (TP), false-positive (FP) and false-negative (FN) rates, as well as in its drift-identification delay.

To assess OPTWIN, we used the popular MOA framework \cite{DBLP:moa} and compared OPTWIN to all major drift-detection frameworks provided in the literature: ADWIN \cite{bifet2007learning}, DDM \cite{gama2004learning}, EDDM \cite{baena2006early}, STEPD \cite{nishida2007detecting} and ECDD \cite{ross2012exponentially}. Using MOA, we compared the {\em precision}, {\em recall} and {\em F1-score}, as well as the {\em delay} of all drift detectors over both sudden and gradual drifts. Furthermore, we also trained a Naive Bayes (NB) classifier on MOA that adapts itself based on the drift detectors and compared its results on various synthetic datasets (STAGGER \cite{schlimmer1986incremental}, RANDOM RBF \cite{bifet2009new}, and AGRAWAL \cite{agrawal1993database}) and real-world ones (Covertype and Electricity) \cite{blackard1999comparative,harries1999splice}. Furthermore, we likewise assessed OPTWIN on a Neural Network (NN) use-case. Specifically, we pre-trained a Convolutional NN (CNN) with the CIFAR-10 \cite{krizhevsky2009learning} image dataset and simulated an end-to-end OL scenario with concept drifts by using OPTWIN and ADWIN as detectors.

Our results show that OPTWIN reliably identifies both sudden and gradual drifts across all datasets. Furthermore, OPTWIN is the drift detector with the best F1-score when compared to the baselines. This is due to its higher precision, which indicates a low FP rate. With respect to the drift-identification delay, there was no drift detector that was superior to the others in a majority of the datasets. In terms of run-time, OPTWIN in combination with the CNN training pipeline is 21\% faster than a similar ADWIN pipeline due to OPTWIN's lower FP rate, which leads to a significantly reduced amount of retraining iterations. Therefore, our experiments indicate that OPTWIN identifies concept drifts with a similar delay as other drift detectors but maintains a higher precision and recall in the drift detection, which reduces the overall run-time of OL pipelines that trigger a re-training of their models upon each detected concept drift.

\section{Background \& Related Works}

We next formally introduce the problem of concept drift and review its principal characteristics (cf. Section \ref{sec:background}). We also discuss the most important related approaches for concept-drift detection from the literature, which serve as inspiration and baselines of our work throughout this paper (cf. Section \ref{sec:related-works}).

\subsection{Background}
\label{sec:background}

A concept drift may exhibit different characteristics whose understanding is essential for a fast and reliable detection. 
Consider an unbounded data stream that receives {\em features} $x_i \in X$ and {\em labels} $y_i \in Y$ in the form of pairs of instances $(x_i,y_i)$. At {\em time} $t$, these instances follow a certain distribution $P_t(X,Y)$. A concept drift is the change of this distribution over time. Formally, a concept drift occurs at time $t+1$ iff $P_t(X,Y) \neq P_{t+1}(X,Y)$ \cite{lu2016concept, lu2018learning, gama2014survey}. 
The underlying data distribution $P_t(X,Y)$ may also be decomposed and expressed as a product of probabilities $P_t(X,Y) = P_t(X) \times P_t(Y|X)$ via the well-known {\em chain rule of probability}. Thus, a concept drift may come from two {\em sources}: (1) $P_t(X) \neq P_{t+1}(X)$, which is known as a {\em virtual drift} because it does not impact the decision boundaries of the learner; and (2) $P_t(Y|X) \neq P_{t+1}(Y|X)$, which is known as an {\em actual drift} because it directly impacts the learner's accuracy \cite{lu2016concept, lu2018learning, gama2014survey}. It is also possible to have both drift sources simultaneously \cite{lu2018learning}.

\begin{figure}[ht]
  \centering
  \vspace{-2mm}
  \includegraphics[width=0.5\textwidth]{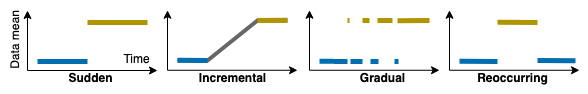}
  \vspace{-2mm}
  \caption{Concept drift types.}
  \label{fig:drift-type}
\end{figure}

Another important characteristic of a concept drift is its {\em type}. Concept drifts can be (1) {\em sudden}; (2) {\em incremental}; (3) {\em gradual}; and (4) {\em reoccurring} (see Figure \ref{fig:drift-type}) \cite{lu2018learning,gama2014survey}. Sudden drifts occur when the probability distribution changes completely within a single step. Incremental drifts occur when the distribution $P_t(X,Y)$ changes incrementally until its convergence. Gradual drifts occur when the new distribution gradually replaces the old one. Reoccurring drifts occur when distributions can reoccur after some time. 

To recover from concept drifts, one may adapt the learner according to two main strategies: (1) {\em active drift adaptation}, which triggers drift adaptation only after a concept drift is flagged by the drift detector; and (2) {\em passive drift adaptation}, in which the learner is constantly adapting itself, independently of the detection of drifts \cite{heusinger2020passive}. The active drift-adaptation strategy depends on the correct identification of the drifts but avoids spending computational resources and time on (re-)training a learner that has already converged. Therefore, we further focus on the active drift-adaptation strategy in the following. 

 There are many sub-strategies for active drift adaptation. The selection of the appropriate one depends not only on the type and source of the drifts but also on the use-case and the adopted learner. A common strategy is to train a new model with the latest data points whenever a drift is identified. Another strategy is to adjust the current learner instead of training a new one from scratch. Moreover, it is also possible to have an ensemble of learners, which is primarily useful for adapting to reoccurring drifts \cite{lu2018learning}. 

The most popular procedure to evaluate learning algorithms that handle concept drifts is the {\em prequential procedure}. Prequential is an evaluation scheme in which each data point is used for testing before training the learning algorithm. Therefore, it is not necessary to know when a drift occurred to perform the evaluation, which is useful when using real-world datasets that do not have labeled drifts \cite{lu2018learning}. Regarding the metrics used to evaluate the drift detector, one may refer to the following ones: TPs, FPs and FNs (and hence {\em precision}, {\em recall} and {\em F1-score}) in the detection rates, as well as the {\em delay} of the detection \cite{lu2018learning, dasu2006information}.

\subsection{Related Works}
\label{sec:related-works}

\noindent\textbf{ADWIN} \cite{bifet2007learning} (for ``ADaptive WINdowing'') is an algorithm with a sliding window $W$ that takes as input a confidence level $\delta \in (0,1)$ and an unbounded sequence of real values $X=\langle x_1, x_2, ..., x_i, ... \rangle$ from a data stream. Each input $x_i$ is expected to be in $[0,1]$ and is stored in a finite window $W\subset X$. The idea of the algorithm is that a concept drift occurs whenever there are two subsequent sub-windows of $W$ with an average difference in their mean above $\epsilon_{cut}$, which is guaranteed to yield a {\em confidence level} of $\delta$. 
If a drift is detected, a part of $W$ is dropped, and the algorithm is ready to resume the drift detection. ADWIN has guaranteed bounds on all error distributions. However, the observed errors usually tend to follow a normal distribution. Thus, they tightened their $\epsilon_{cut}$ accordingly and evaluated the algorithm taking those new bounds. Furthermore, as ADWIN checks multiple sub-windows of $W$, its complexity is $\mathcal{O}(\log{}|W|)$ per iteration (where an iteration is ideally triggered for each new element that arrives from the stream).

\noindent\textbf{DDM} \cite{gama2004learning} (for ``Drift Detection Method'') also receives as parameter a value $\delta$ related to the confidence level of the algorithm, and a sequence of data instances $X$ which are available one by one from a data stream. However, the data stream is expected to follow a binomial distribution, as its values represent the number of errors in a sample of $n$ instances. Therefore, it is primarily suited for classification problems rather than for regression. The main idea of DDM is that, as the number of samples increases, the number of errors from a learner will decrease. Therefore, if the error rate from the learner increases above a certain threshold, it means that a concept drift occurred. Thus, DDM tracks the minimal error rate $p_{min}$ from $X$ and its respective standard deviation $s_{min}$. Then, DDM flags a drift at step $i \geq 30$ iff $p_i + s_i \geq p_{min} + \delta s_{min}$. When a drift is detected, the algorithm is reset. One advantage of DDM is its lightweight nature due to storing only $p_{min}$ and $s_{min}$. However, DDM is known for its decreased performance when detecting slow gradual changes. 

\noindent\textbf{EDDM} \cite{baena2006early} (for ``Early Drift Detection Method'') is similar to DDM but uses the distance between errors to identify concept drifts. Therefore, EDDM is suited only for binary input data, which does not allow it to be used for regression problems. They assume that, as the number of examples increases, the distance between errors will increase accordingly. Therefore, if this distance decreases more than a threshold, a concept drift is flagged. Thus, EDDM tracks the maximum distance between errors $p_{max}'$ and its respective standard deviation $s_{max}'$. Then, after analyzing at least 30 errors, EDDM flags a drift at step $t$ if $(p_{t}' + 2\,s_{t}') / (p_{max}' + 2\,s_{max}' < \delta)$, with $\delta$ again representing a simple form of confidence level for the detected concept drifts. According to the authors, EDDM improves DDM's performance when identifying gradual concept drifts while maintaining good performance on sudden drifts. 

\noindent\textbf{STEPD} \cite{nishida2007detecting} (for using the ``Statistical Test of Equal Proportions'') is a lightweight drift detector that also considers a sequence of data instances $X$ as input. The idea of STEPD is that the recent accuracy of a learner shall be similar to its overall accuracy, otherwise a concept drift occurred. As the name suggests, STEPD checks the equality-of-proportions hypothesis test \cite{massey2006tests} by comparing the most recent values from $X$ (kept in a sliding window $W$) with the other values in $X$, thus comparing $X_{0:W}$ and $X_{W:|X|-1}$. It also considers that the data follows a normal distribution, and with a confidence level of $\delta$, it can determine if the null hypothesis is rejected and a concept drift is flagged. In this case, the algorithm is reset. The value of $W$ selected by the authors was 30. 

\noindent\textbf{ECDD} \cite{ross2012exponentially} is the acronym for ``Exponentially Weighted Moving Average'' (EWMA) for concept-drift detection. It is a drift-detection method based on EWMA that checks the misclassification rate of its underlying learners. ECDD receives a sequence of data instances $X$ as input, which it assumes to follow a Bernoulli distribution. ECDD is suited only for classification problems, as it only accepts binary data from the data stream. Other than $X$, ECDD takes as inputs $ARL_0$, which is the time one expects between detecting false positives; and $\lambda$, which dictates how much weight new data has compared to the old one. Despite being computationally expensive to calculate, $ARL_0$ can be approximated before starting to process the data stream. ECDD then uses pre-calculated polynomials, $ARL_0$ and $\lambda$ to update its estimators. With its updated estimators, ECDD detects if a concept drift occurred or not. Finally, if a drift is detected, the algorithm is reset. 

We note that all algorithms described above additionally have a form of warning detection which can be used to assist in the concept drift adaptation. The warning is usually a simple relaxation of the drift threshold which can be used to start training a new learner before the actual concept drift is flagged. 

\section{OPTWIN -- Optimal Window Concept Drift Detector}
OPTWIN is the new concept-drift detector that we propose in this paper. It is an error rate-based drift detector, which tracks the error rates produced by an OL learner in a sliding window $W$. Its name stands for ``OPTimal WINdow'' due to its calculation of the optimal split of the sliding window. We consider this split ``optimal'' because it is detected for the first element in $W$ at which a statistically significant difference between either the means or the variances of the error rates (based on the common $t$- and the $f$-tests) among two sub-windows of $W$ occurs.
Specifically, $W$ keeps growing until either a concept drift is detected or a maximum user-defined size $w_{max}$ is reached. Based on the sliding-window size $|W|$ and a pre-defined confidence level $\delta$, OPTWIN calculates the optimal point to divide $W$ into ``historical'' ($W_{hist}$) and ``new'' ($W_{new}$) data points. 

\subsection{Definition of concepts}
\squishlist
\item $\delta$: confidence level defined by the user, $\delta \in (0,1)$.
\item $W$: sliding window of error rates.
\item $w_{proof}$: minimum size of $W$ to identify drifts of $\rho$ magnitude.
\item $w_{min}$: minimum $W$ size, with $w_{min} = 30$.
\item $w_{max}$: maximum $W$ size, with $w_{max} \in [w_{min}, \infty)$.
\item $\mu_W$: mean of $W$.
\item $\sigma_W$: standard deviation of $W$.
\item $\nu$: optimal splitting percentage of $W$, with $\nu \in (0,1)$. 
\item $\nu_{split}$: optimal splitting point of $W$, with $\nu_{split} = \lfloor\nu\,|W|\rfloor$ 
\item $W_{hist}$: historical errors from $W$, defined as $W_{0\,:\,\nu_{split}}$.
\item $W_{new}$: new errors from $W$, defined as $W_{\nu_{split}\,:\,|W|-1}$.
\item $\rho$: robustness parameter, i.e., the ratio that $\mu_{W_{new}}$ has to vary in relation to $\sigma_{W_{hist}}$ to count as a concept drift, with $\rho \ in (0, \infty)$. 
\item $\rho_{temp}$: temporary $\rho$, defined by solving Equation \ref{eq:nu}, with $\nu = 0.5$.
\item $X={x_1, x_2, ..., x_i, ... }$: data-stream of real numbers $x_i$. 
\squishend

\subsection{Setup \& Parameters}
Without loss of generality, we consider a typical OL scenario, in which OPTWIN receives as input a sequence of real numbers $x_1, x_2, ..., x_i, ... $ from a possibly unbounded data-stream $X$. Thus, the algorithm is assumed to access only one element $x_i$ at time $i$ from the stream, and buffers previously seen elements in a {\em sliding window} $W \subset X$ of consecutive events. Moreover, OPTWIN requires as parameters (1) $\delta$, the {\em confidence level} for the concept-drift detection; (2) $w_{max}$, the {\em maximum size} of the sliding window $W$; and (3) $\rho$, a parameter we refer to as {\em robustness}, which we define as the minimum ratio by which $\mu_{W_{new}}$ has to vary in relation to $\sigma_{W_{hist}}$ in order for OPTWIN to consider this variation as a concept drift.


\subsubsection{Assumptions}
\squishlist
\item There is no concept drift within the first $w_{min}$ data instances from $X$ (which is needed to initialize OPTWIN). 
\item A concept drift occurs when the means or the standard deviations within two sub-windows of $W$ are statistically different.
\item Within any two sub-windows of $W$, the values produced by the test statistic of the unequal-variance $t$-test \cite{ruxton2006unequal} (henceforth called ``$t\_value$'') follow a $t$-distribution.
\item Within any two sub-windows of $W$, the values produced by the test statistic of the $f$-test \cite{mahbobi2016introductory} (henceforth called ``$f\_value$'') follow an $f$-distribution.
\squishend

Our first assumption mitigates the negative impact of outliers when a few data points are available. The second assumption points out when we expect OPTWIN to identify concept drifts. The third and the fourth assumptions are generally required to guarantee the validity of the $t$- and $f$-tests, respectively, but do not impose limitations on the values ingested from the data stream in practice. 

\subsection{Algorithm}
\label{sec:alg}

{\small
\begin{algorithm}
\caption{OPTWIN}\label{alg}
\begin{algorithmic}[1]
\Statex \textbf{Input parameters:} \hfill \textbf{Global variables:} 
\Statex \textbullet~$\delta$ -- confidence level \hfill $W = \langle\,\rangle$ -- sliding window
\Statex \textbullet~$\rho$ -- robustness \hfill $w_{min} = 30$  -- min window size
\Statex \textbullet~$w_{max}$-- max window size \hfill $\eta=1e^{-5}$ -- avoids division by 0
\Statex
\Procedure{AddElement}{$x_i$}
\State $W \gets W \cup  x_i$
\If {$|W| < w_{min}$}
    \State 
    \Return False
\ElsIf {$|W| \geq max\_lenght$}
\State $W \gets W - W_0$
\EndIf
\State $\nu \gets \textsc{OptimalCut}(|W|, \rho, \delta^{\frac{1}{4}}\textit{)}$ \hfill cf. Equation \eqref{eq:nu}
\State $\nu_{split} \gets \lfloor\nu\,|W|\rfloor$ 
\State $W_{hist} \gets W_{0:\nu_{split}}$
\State $W_{new} \gets W_{\nu_{split}:|W|-1}$
\Statex \hfill// \textit{f-test}
\If {$\frac{(\sigma_{W_{new}}+\eta)^2}{(\sigma_{W_{hist}}+\eta)^2} > f\_ppf(\delta^{\frac{1}{4}}, \nu|W|-1, (1-\nu)|W|-1)$}
    \State reset()
    \State
    \Return True
\Statex \hfill// \textit{t-test} -- cf. Equations \eqref{eq:nu} and \eqref{eq:df}
\ElsIf {$t\_value(W_{hist}, W_{new}) > t\_ppf(\delta^{\frac{1}{4}}, df)$}
    \State reset()
    \State 
    \Return True
\EndIf
\EndProcedure
\end{algorithmic}
\end{algorithm}
}

As seen in Algorithm \ref{alg}, OPTWIN first needs to be initialized by creating an empty sliding window $W$. Analogously to other drift detectors, it collects $w_{min}$ many first elements from the stream (usually within 30 to 50) 
to guarantee that it has a minimum amount of data to output statistically relevant drift detections. It also defines a small constant $\eta = 1e^{-5}$, which avoids a division by 0 when added to the standard deviations during the calculation of the $f$-test. 

Second, the $\textsc{AddElement}$ procedure is called once to process each data element received from the data stream. The procedure starts by inserting the most recent data element $x_i$ into $W$ and by checking whether $W$ has already reached the minimum amount of elements to detect a possible concept drift. If so, it also checks if $W$ increased above $w_{max}$ and removes the oldest element from $W$ if necessary. Up to this point, OPTWIN performs standard operations to maintain its sliding window bounded between $w_{min}$ and $w_{max}$ elements.

Then, OPTWIN calculates $\nu$, which is the optimal splitting point of $W$ into $W_{hist}$ and $W_{new}$, by solving Equation \ref{eq:nu} for the highest value of $\nu$ in terms of $\delta'$, $\rho$ and $|W|$:
\begin{equation}
\label{eq:nu}
\rho = t\_ppf(\delta', df) \sqrt{\frac{1}{\nu|W|}+\frac{f\_ppf(\delta', \nu\,|W|-1, (1-\nu)\,|W|-1)}{(1-\nu)\,|W|}}
\end{equation}
where $\rho$ is the aforementioned, user-defined robustness parameter.
During this calculation, OPTWIN uses $t\_ppf$ and $f\_ppf$, which are the Probability Point Functions (PPF) of the $t$- and the $f$-distributions, respectively\footnote{We omit the arguments of $f\_ppf$ in the following equation to improve readability.}.

To calculate $t\_ppf$, we apply a confidence of $\delta' = \delta^{\frac{1}{4}}$, which considers the application of the two tests to calculate $\nu$ in Equation \ref{eq:nu} and the two tests on Lines 11 and 14 in Algorithm \ref{alg}. Equation \ref{eq:df} depicts our calculation of the degrees of freedom $df$, which is needed as another parameter for the $t$-distribution:
\begin{equation}
\label{eq:df}
df = \frac{(\frac{1}{\nu|W|}+\frac{f\_ppf}{(1-\nu)|W|})^2}{\frac{(\frac{1}{\nu|W|})^2}{(\nu|W|)-1}+\frac{(\frac{f\_ppf}{(1-\nu)|W|})^2}{((1-\nu)|W|)-1}}
\end{equation}

To calculate $f\_ppf$, on the other hand, we take $\nu|W| - 1$ and $(1-\nu)|W| - 1$ as the degrees of freedom for the $f$-distribution and again apply the same confidence value $\delta'$ also here.

Finally, OPTWIN performs the $t$- and $f$-tests (in any order) to compare both sub-windows and determine if their means and standard deviations, respectively, belong to the same distribution. If not, a concept drift is flagged and the algorithm is reset. Otherwise, the $\textsc{AddElement}$ method is called for the next element from the data stream in an iterative manner.

The usage of the $t$- and $f$-tests to identify significant changes in the means and standard deviations among series of data values is well-established. However, OPTWIN's novelty comes from the combination of both tests to calculate $\nu$, and thereby identify where to optimally divide this series of values to perform both tests. In short, by solving Equation \ref{eq:nu} in terms of $\nu$, we determine the minimum size of $W_{new}$ (the optimal splitting point of $W$) that statistically guarantees the identification of any concept drift with a robustness of at least $\rho$ when using the $t$- and $f$-tests, thereby achieving lower drift-detection delays than other approaches. To calculate $\nu$, other than the confidence level $\delta$, the only values that shall be inputted by the user are $w_{max}$ and $\rho$. Regarding $\rho$, when selecting a small value, one may expect to identify smaller drifts in exchange of a higher drift-detection delay. On the other hand, with higher values of $\rho$, one can expect a smaller detection delay in exchange of missing smaller drifts. In practice, $\rho$ shall be set based on the magnitude and frequency of drifts expected. However, determining the correct $\rho$ to each use case should not be a difficult task, since different $\rho$'s tend to produce similar results (as seen in Section \ref{sec:experiments}). 
Regarding $w_{max}$, with a higher value, one may expect smaller drift-detection delays in exchange for more memory usage. In practice, when using $\rho = 0.1$, we did not observe a significant variation in $|W_{new}|$ even when increasing $w_{max}$ to more than 25,000 elements.

One important point to note is that we can only calculate $\nu$ as the optimal splitting point if $W$ is large enough. Otherwise, we set $\nu$ to the middle of $W$ until it grows to the minimum size needed to solve Equation \ref{eq:nu}. Thus, if it exists, it is defined as the highest root of Equation \ref{eq:nu}. Otherwise, it is set to $\nu = 0.5$.


Below, we present Theorem \ref{theorem}, our main theoretical result. It gives guarantees in terms of OPTWIN's false positive (FP) and false negative (FN) bounds for identifying concept drifts.

\begin{theorem}
\label{theorem}
\item
\squishlist

\item\textbf{False Positive Bound.} At every step, if $\mu_W$ and $\sigma^2_W$ remain constant within $W$, OPTWIN will flag a concept drift at this step with a confidence of at most 1-$\delta$. 

\smallskip
\item\textbf{False Negative Bound} (for mean drift with large enough $W$). For any partitioning of $W$ into two sub-windows $W_{hist}\,W_{new}$, with $|W|\geq w_{proof}$ and $W_{new}$ containing the most recent elements, if $\mu_{hist}-\mu_{new} > \rho\,\sigma_{hist}$, then, with confidence $\delta$, OPTWIN flags a concept drift in at most $|W| - \nu_{split}$ steps.

\smallskip
\item \textbf{False Negative Bound} (for mean drift with small $W$). For any partitioning of $W$ into two sub-windows $W_{hist}\,W_{new}$, with $w_{min} \leq |W| < w_{proof}$ and $W_{new}$ containing the most recent elements, if $\mu_{hist}-\mu_{new} > \rho_{temp}\,\sigma_{hist}$, then, with confidence $\delta$, OPTWIN flags a concept drift in at most $\frac{|W|}{2}$ steps.

\smallskip
\item \textbf{False Negative Bound} (for standard-deviation drift with any $W$). For any partitioning of $W$ into two sub-windows $W_{hist}\,W_{new}$, with $|W| \geq w_{min}$ and $W_{new}$ containing the most recent elements, if $\frac{\sigma^2_{new}}{\sigma^2_{hist}} > f\_ppf(\delta', \nu\,|W|-1, (1-\nu)\,|W|-1)$, then, with confidence $\delta$, OPTWIN flags a concept drift in at most $\nu_{split}$ steps. 
\squishend
\end{theorem}

\subsubsection{Proof of Theorem \ref{theorem}}

\squishlist
\item\textbf{Part 1:} In each iteration of OPTWIN, we apply the $t$- and the $f$-tests to identify the concept drifts. Therefore, the false-positive bounds of OPTWIN are directly drawn from those tests. Thus, with $\delta' = \delta^\frac{1}{4}$ as the confidence for each of the four tests, we identify a false-positive drift with a confidence of at most $1-\delta$.

\smallskip
\item\textbf{Part 2:} Considering an unequal-variance $t$-test \cite{ruxton2006unequal} by comparing $W_{hist}$ and $W_{new}$, we have:
\begin{equation}
t\_value = \frac{\mu_{hist} - \mu_{new}}{\sqrt{\frac{\sigma_{hist}^2}{|W_{hist}|}+\frac{\sigma_{new}^2}{|W_{new}|}}}
\end{equation}

The nominator on the right represents the difference observed between the means of $W_{hist}$ and $W_{new}$. If this difference is large enough, we can statistically guarantee that both sets have distributions with different means. To simplify the equation, we represent this difference in terms of $\sigma_{hist}$. Therefore, we set $\mu_{hist} - \mu_{new} = \rho\,\sigma_{hist}$ with $\rho$ being a user-defined robustness parameter. Thus, by applying this substitution and passing the denominator to the other side of the equation, we have:
\begin{equation}
\rho\,\sigma_{hist} = t\_value\sqrt{\frac{\sigma_{hist}^2}{|W_{hist}|}+\frac{\sigma_{new}^2}{|W_{new}|}}
\end{equation}

Considering $\nu$ as the splitting point of $W$ into $W_{hist}$ and $W_{new}$, we have $|W| = \nu\,|W_{hist}| + (1-\nu)\,|W_{new}|$, and therefore:
\begin{equation}
\label{eq:middle}
\rho\,\sigma_{hist} = t\_value\sqrt{\frac{\sigma_{hist}^2}{\nu\,|W|}+\frac{\sigma_{new}^2}{(1-\nu)\,|W|}}
\end{equation}

By our definition of a concept drift, $\sigma_{new}$ and $\sigma_{hist}$ must follow the same distribution (otherwise a drift occurred). Furthermore, we also consider that they follow the $f$-distribution, which is usually the case for standard deviations. Thus, consider the definition of the test statistic for the $f$-test \cite{mahbobi2016introductory} as shown below:
\begin{equation}
\label{eq:f-test}
f\_value = \frac{\sigma_{new}^2}{\sigma_{hist}^2}
\end{equation}

We can now determine $\sigma_{new}$'s upper bound in terms of $\sigma_{hist}$ and by defining $f\_factor$ as the highest value that the $f\_value$ can have while maintaining the null hypothesis of the $f$-test with confidence $\delta'$. 
\begin{equation}
\label{eq:f-factor}
\sigma_{hist}^2\,f\_factor \geq \, \sigma_{new}^2
\end{equation}

In this case, the $f\_factor$ can be calculated using the $f\_ppf$, the PPF of the $F$-curve, by taking a confidence of $\delta'$ and $|W_{new}|-1$ and $|W_{hist}|-1$ as the degrees of freedom for the numerator and the denominator, respectively (cf. Equation \ref{eq:f-test}). Thus, we obtain:
\begin{equation}
f\_factor = f\_ppf(\delta', \nu\,|W|-1, (1-\nu)\,|W|-1)
\end{equation}

Therefore, to later simplify Equation \ref{eq:middle}, we substitute $\sigma_{new}$ with its upper bound as calculated by the $f$-test on Equation \ref{eq:f-factor}:
\begin{equation}
\rho\,\sigma_{hist} = t\_value\sqrt{\frac{\sigma_{hist}^2}{\nu\,|W|}+\frac{\sigma_{hist}^2 \, f\_factor}{(1-\nu)\,|W|}}
\end{equation}

Next, we can simplify the equation by removing $\sigma_{hist}$ from both sides as follows:
\begin{equation}
\label{eq:final}
\rho = t\_value\sqrt{\frac{1}{\nu\,|W|}+\frac{f\_factor}{(1-\nu)\,|W|}}
\end{equation}

Then, we calculate the $t\_value$ in Equation \ref{eq:t-value} analogously by using the PPF of the $T$-curve. 
\begin{equation}
\label{eq:t-value}
t\_value = t\_ppf(\delta', df)
\end{equation}

We consider a confidence level of $\delta'$ and the degrees of freedom $df$ for the unequal-variance $t$-test (cf. Equation \ref{eq:df1}).
\begin{equation}
\label{eq:df1}
df = \frac{(\frac{\sigma_{hist}^2}{|W_{hist}|}+\frac{\sigma_{new}^2}{|W_{new}|})^2}{\frac{(\frac{\sigma_{new}^2}{|W_{hist}|})^2}{|W_{hist}|-1}+\frac{(\frac{\sigma_{new}^2}{|W_{new}|})^2}{(|W_{new}|)-1}}
\end{equation}

By considering Equation \ref{eq:df1} in our setting with $|W_{hist}| = \nu\,|W|$ and $|W_{new}| = (1-\nu)\,|W|$ and our upper bound for $\sigma_{new} \leq \sigma_{hist}\,f\_factor$ (cf. Equation \ref{eq:f-factor}), we reach the formerly presented Equation \ref{eq:df}.

Finally, by replacing $t\_value$ on Equation \ref{eq:final} by the one calculated on Equation \ref{eq:t-value}, we reach Equation \ref{eq:nu} introduced in Section \ref{sec:alg}.\qed
\squishend

We remark that Equation \ref{eq:nu} only depends on $\delta'$, $\rho$, $|W|$ and $\nu$. Recall that $\delta'$ is $\delta^{\frac{1}{4}}$ which is provided by the user; $\rho$ is also given by the user; and $|W|$ increases linearly until $w_{max}$. We can then calculate the highest value $\nu$ based on each $|W|$ that solves Equation \ref{eq:nu}. 

Therefore, considering the split $W = W_{hist}\,W_{new}$ with $W_{hist} = W_{0:\nu\,|W|}$ and $W_{new} = W_{\nu\,|W|:|W|-1}$, by applying the $t$-test, we guarantee that if $\mu_{hist}$ and $\mu_{new}$ diverge more than $\rho\,\sigma_{hist}$, we identify this divergence with a confidence of $\delta$. Moreover, considering our OL setting, it takes no more than $(1-\nu)|W|$ iterations for $W$ to be divided into $W_{hist}\,W_{new}$ after the drift occurred. 

\squishlist
\item\textbf{Part 3:} By setting $\nu = 0.5$ in Equation \ref{eq:final}, we have:
\begin{equation}
\rho_{temp} = t\_value\sqrt{\frac{1}{0.5\,|W|}+\frac{f\_factor}{0.5\,|W|}}
\end{equation}

With $\nu = 0.5$ and $\rho_{temp}$ as calculated above, we can guarantee that if $\mu_{hist}$ and $\mu_{new}$ diverge more than $\rho_{temp}\,\sigma_{hist}$, we identify the divergence with a confidence of $\delta$ (as guaranteed by the test procedure). Similarly to Part 2 of the proof, we achieve this in at most $(1-\nu)\,|W|$ iterations, in this case, $\frac{|W|}{2}$ iterations.\qed

\smallskip
\item\textbf{Part 4:} This bound is directly obtained from the $f$-test. By simply applying the $f$-test to compare two sub-windows of $W$, we guarantee the given confidence value, in this case $\delta$. Again, it takes no more than $(1-\nu)\,|W|$ iterations for $W$ to be divided into $W_{hist}\,W_{new}$ after the drift occurred.\qed
\squishend

\subsection{Implementation \& Analysis}
\label{sec:implementation}

We implemented OPTWIN via a combination of Java and Python scripts as extensions of the MOA \cite{DBLP:moa} (Java) and the River \cite{montiel2021river} (Python) libraries. We pre-calculated the values of $\nu$, $t\_ppf$, and $f\_ppf$ based on a fixed confidence value of $\delta=0.99$ and for $w_{max} = 25,000$, thus $30 \leq |W| \leq$ 25,000. All pre-calculated variables were stored in lists indexed by the $|W|$ from which they were calculated. This is possible because those variables, as seen in Equation \ref{eq:nu}, do not depend on the data distribution. Thus, it is not necessary to calculate them in real-time. On the other hand, we need to store the sliding window $W$ plus the other three lists of floating point values of size up to $w_{max}$ in memory, which takes a maximum of $w_{max}*4*4$ bytes. Thus, for $w_{max} =$ 25,000, OPTWIN would require only around 390 KB of memory. 

Furthermore, the full calculation of the means and standard deviations from $W_{hist}$ and $W_{new}$ can be avoided. Instead of calculating them from scratch, one only needs to update them incrementally. Moreover, as $W$ is bounded by $w_{max}$, we can use a circular array to make insertions at the end of the array, deletions from the beginning of the array, and then look up each value in $\mathcal{O}(1)$ time. Therefore, the $\textsc{AddElement}$ procedure has an overall computational complexity of $\mathcal{O}(1)$ per element ingested from the data stream (assuming a constant cost for the numerical operations involved in resolving Equation \ref{eq:nu} to $\nu$). Our analytical approach thus provides great potential runtime gains over the iterative search procedure applied, e.g., by ADWIN which requires $\mathcal{O}(\log_{}|W|)$ computations per iteration. 

By default, OPTWIN's Algorithm \ref{alg} tracks concept drifts that either increase or decrease the means and standard deviations of the variables tracked. However, in an OL scenario, we usually want to update the learner only when the number of errors increases. Therefore, in our implementation, we check if also $\mu_{new} \geq \mu_{hist}$ along with the statistical tests on Lines 11 and 14 of Algorithm \ref{alg}. In doing so, we consider that a drift occurred only if $\mu_{new}$ is higher than $\mu_{hist}$ (i.e., the learner decreased in performance). 
\section{Experiments \& Results}
\label{sec:experiments}

In this section, we report our detailed experiments to asses OPTWIN. We compared OPTWIN to the most-commonly used baselines for concept-drift detection: ADWIN, DDM, EDDM, STEPD, and ECDD. We performed most of the experiments using the common MOA \cite{DBLP:moa} framework which is a Java-based data stream simulator. In addition, we performed experiments on Python to asses OPTWIN on a Neural Network (NN) use case, which would not be possible via MOA. 

\subsection{MOA Experiments}
We compared 3 configurations of OPTWIN with the default configurations of our baselines. All OPTWIN configurations had $\delta=0.99$, $w_{max} =$ 25,000, and pre-computed values for $\nu$, $t\_ppf$, and $f\_ppf$ (as described on Section \ref{sec:implementation}). The difference among the OPTWIN configurations is only on $\rho$, which we varied between $0.1$, $0.5$, and $1.0$ to better understand how our robustness parameter affects OPTWIN's performance in practice.

We performed two types of experiments on MOA. The first one uses the ``Concept Drift'' interface, in which MOA creates a stream of data points (either binary or non-binary) and produces a concept drift that can be sudden or gradual. We later refer to those experiments according to their data input and their drift type.

The second type of experiment uses the ``Classification'' interface. It generates data streams based on synthetic datasets (STAGGER, RANDOM RBF, and AGRAWL) \cite{schlimmer1986incremental,bifet2009new,agrawal1993database} and real-world ones (Electricity and Covertype) \cite{blackard1999comparative,harries1999splice}. The idea of these experiments is to train a classifier that is reset every time a concept drift is detected by a drift detector. We chose MOA's built-in Naive Bayes (NB) classifier for its simplicity, which facilitates the analysis of the drift-detection results. For the synthetic datasets, we generate data streams with 100,000 data points with drifts occurring every 20,000 data points (either sudden or gradual). For the real-world data sets, the drifts are already present and have an unknown location on the stream.

Table \ref{tb:false_positive} presents the comparison among drift detectors on the above-mentioned configurations. We repeated each experiment 30 times and compared their average TP, FP and FN rates to compute their micro-average precision, recall, and F1-score, along with their average drift-detection delay (in terms of the number of streamed elements between the occurrence of a known concept drift and its identification by the detector). Note that we did not include in Table \ref{tb:false_positive} the ``Classification'' experiments on real-world datasets nor the ones with gradual concept drifts. For the real-world datasets, it is not possible to calculate the above-mentioned metrics without knowing the drift's locations. For the gradual drifts, we observed a divergence between the starting and ending points of those drifts in the MOA documentation and in practice. Therefore, we did not include those configurations in our comparison. 

{
\setlength{\abovecaptionskip}{-12pt}
\small
\begin{table}
\centering
\begin{tabular}{l l c c c c c}
\hline
\textbf{Experiment} & \textbf{Drift Detector} & \textbf{Delay} & \textbf{FP} & \textbf{P} & \textbf{R} & \textbf{F1} \\
& ADWIN & 280 & 16.33 & 43\% & 100\% & 60\% \\
& DDM & 365 & 0.37 & 88\% & 100\% & 93\%\\
gradual & EDDM & 148 & 6.33 & 56\% & 100\% & 71\% \\
binary & STEPD & 180 & 8.00 & 23\% & 100\% & 37\% \\
drift & ECDD & \textbf{117} & 5.37 & 27\% & 100\% & 42\% \\
& OPTWIN $\rho=0.1$ & 328 & 0.30 & 88\% & 100\% & 94\% \\
& OPTWIN $\rho=0.5$ & 278 & 0.10 & 96\% & 100\% & 98\% \\
& OPTWIN $\rho=1.0$ & 317 & 0.07 & 97\% & 100\% & \textbf{99}\% \\
\hline
& ADWIN & 145 & 0.00 & 100\% & 100\% & \textbf{100}\% \\
gradual & STEPD & 41 & 150 & 10\% & 100\% & 18\% \\
non-binary & OPTWIN $\rho=0.1$ & \textbf{2.00} & 0.00 & 100\% & 100\% & \textbf{100}\% \\
drift& OPTWIN $\rho=0.5$ & \textbf{2.00} & 0.00 & 100\% & 100\%  & \textbf{100}\% \\
& OPTWIN $\rho=1.0$ & \textbf{2.00} & 0.00 & 100\% & 100\% & \textbf{100}\% \\
\hline
& ADWIN & 46.33 & 1.87 & 35\% & 100\% & 52\% \\
& DDM & 64.50 & 0.20 & 83\% & 100 & 91\%\\
sudden & EDDM & 26.77 & 6.57 & 13\% & 100\% & 23\% \\
binary & STEPD & \textbf{0.40} & 6.67 & 71\% & 100\% & 83\% \\
drift & ECDD & 5.47 & 2.67 & 27\% & 97\% & 42\% \\
& OPTWIN $\rho=0.1$ & 75.13 & 0.00 & 100\% & 100\% & \textbf{100}\% \\
& OPTWIN $\rho=0.5$ & 28.17 & 0.00 & 100\% & 100\% & \textbf{100}\% \\
& OPTWIN $\rho=1.0$ & 18.33 & 0.33 & 75\% & 100\% & 86\% \\
\hline
& ADWIN & 25.00 & 2.00 & 33\% & 100\% & 50\% \\
sudden & STEPD & 21.00 & 32.00 & 3\% & 100\% & 6\% \\
non-binary & OPTWIN $\rho=0.1$ & \textbf{1.00} & 0.00 & 100\% & 100\% & \textbf{100}\% \\
drift& OPTWIN $\rho=0.5$ & \textbf{1.00} & 0.00 & 100\% & 100\%  & \textbf{100}\% \\
& OPTWIN $\rho=1.0$ & \textbf{1.00} & 0.00 & 100\% & 100\% & \textbf{100}\% \\
\hline
& ADWIN & 31.00 & 0.30 & 93\% & 100\% & 96\% \\
& DDM & 6.33 & 0.17 & 96\% & 100\% & 98\% \\
sudden & EDDM & 40.01 & 0.00 & 100\% & 100\% & \textbf{100}\% \\
STAGGER & STEPD & 17.39 & 31.40 & 11\% & 99\% & 20\% \\
& ECDD & 0.72 & 0.47 & 90\% & 100\% & 94\% \\
& OPTWIN $\rho=0.1$ & 0.76 & 0.27 & 94\% & 100\% & 97\% \\
& OPTWIN $\rho=0.5$ & \textbf{0.72} & 0.43 & 90\% & 100\% & 95\% \\
& OPTWIN $\rho=1.0$ & \textbf{0.72} & 0.60 & 87\% & 100\% & 93\% \\
\hline
& ADWIN & 169.67 & 6.00 & 33\% & 75\% & 46\%\\\
& DDM & 536.33 & 2.00 & 60\% & 75\% & 67\% \\
sudden & EDDM & \textbf{53.55} & 11.00 & 21\% & 75\% & 33\% \\
RANDOM & STEPD & 281.25 & 25.00 & 14\% & 100\% & 24\% \\
RBF & ECDD & 71.75 & 174.00 & 2\% & 100\% & 4\% \\
& OPTWIN $\rho=0.1$ & 315.00 & 0.00 & 100\% & 75\% & \textbf{86}\% \\
& OPTWIN $\rho=0.5$ & 187.00 & 0.00 & 100\% & 75\% & \textbf{86}\% \\
& OPTWIN $\rho=1.0$ & 1,931.50 & 1.00 & 67\% & 50\% & 57\% \\
\hline
& ADWIN & \textbf{229.40} & 4.23 & 49\% & 100\% & 65\% \\
& DDM & 1,875.87 & 0.90 & 78\% & 80\% & 79\% \\
sudden & EDDM & 5,370.47 & 17.20 & 12\% & 60\% & 20\% \\
AGRAWAL & STEPD & 838.03 & 23.47 & 14\% & 99\% & 25\% \\
& ECDD & 465.90 & 153.57 & 3\% & 100\% & 5\% \\
& OPTWIN $\rho=0.1$ & 371.62 & 0.63 & 86\% & 100\% & \textbf{93}\% \\
& OPTWIN $\rho=0.5$ & 350.89 & 0.77 & 83\% & 93\% & 88\% \\
& OPTWIN $\rho=1.0$ & 630.63 & 0.27 & 93\% & 88\% & 91\% \\
\hline
\end{tabular}
\caption{Statistics of drift identification on synthetic datasets.}
\label{tb:false_positive}
\end{table}
}

Based on the results in Table \ref{tb:false_positive}, we can generally observe a high F1-score for OPTWIN when compared to the baselines. In fact, with $\rho \leq 0.5$, OPTWIN's average F1-score is over 95\%, followed by OPTWIN with $\rho=1.0$ achieving 89,2\%, DDM with 89,7\%, ADWIN reaching 67\% and the other detectors less than 50\%. We assert this due to OPTWIN's low FP rate which produces a high precision and F1-score, respectively. We further compared the F1-scores of all configurations of OPTWIN with the two drift detectors that can be used for regression problems (ADWIN and STEPD), which showed OPTWIN to be superior based on a one-tailed Wilcoxon signed-rank test with $\alpha=0.05$ in a statistically significant manner. 

Regarding the drift-detection delay, OPTWIN with $\rho = 0.5$ is the drift detector with the smallest delay, taking on average 121 iterations to detect a concept drift; followed by ADWIN, ECDD, and OPTWIN at $\rho = 0.1$ taking 132, 132 and 156 iterations each; the other detectors took on average between 198 and 1127 iterations to detect a drift, with OPTWIN at $\rho = 1.0$ taking 414 on average. Observe that, the higher the FP rate of a drift detector, the higher are also its chances of identifying a drift earlier. Therefore, when analyzing ECDD's average drift delay of 132 iterations, we have to take into consideration that it had an average of 67 FPs per run, in contrast to an average of less than 0.4 for any of the OPTWIN configurations. Furthermore, considering that DDM, EDDM, and ECDD depend on binary input data, they were not included in the experiments using the ``non-binary'' datasets.

We can better visualize the difference between the drift detectors in Figures \ref{fig:abrupt} and \ref{fig:gradual}, which represent one of the 30 runs of the ``sudden binary drift'' and the ``gradual binary drift'' configurations, respectively (we selected the runs with results closest to the ones on the observed average). In Figure \ref{fig:abrupt}, we can see the relatively high FP rate of the EDDM, ADWIN and ECDD detectors when compared to OPTWIN, DDM and STEPD. Moreover, we can see that OPTWIN's detection delay decreases as $\rho$ increases. In Figure \ref{fig:gradual}, we cannot observe the same effect of $\rho$ in OPTWIN's delay detection. Nevertheless, we can still observe the high FP rate of EDDM, ECDD and ADWIN compared to the other drift detectors. We note that STEPD's high FP rate comes from a few runs with dozens of FPs.

{
\setlength{\belowcaptionskip}{-15pt}
\setlength{\abovecaptionskip}{-10pt}
\begin{figure}
\setlength{\columnsep}{0.1\columnsep}
\setlength{\subfigtopskip}{0.1\subfigtopskip}
\setlength{\subfigbottomskip}{0.1\subfigbottomskip}
\setlength{\subfigcapskip}{0.1\subfigcapskip}
\begin{multicols}{2}
    \subfigure{\includegraphics[width=1\columnwidth]{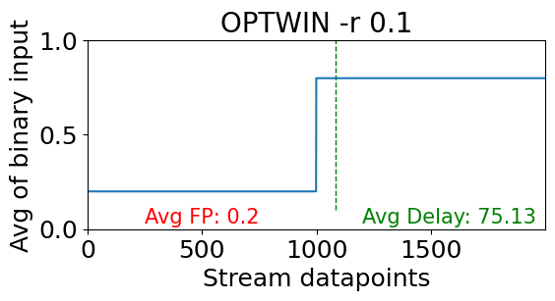}} \\
    \subfigure{\includegraphics[width=1\columnwidth]{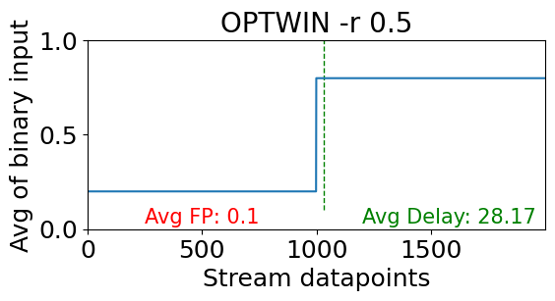}} \\
    \subfigure{\includegraphics[width=1\columnwidth]{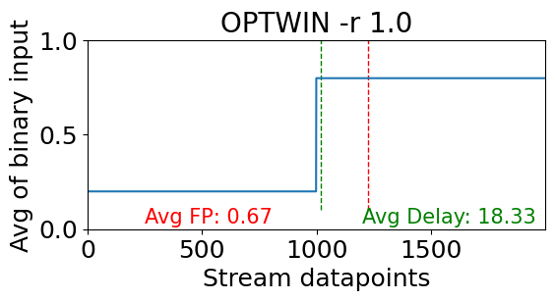}} \\
    \subfigure{\includegraphics[width=1\columnwidth]{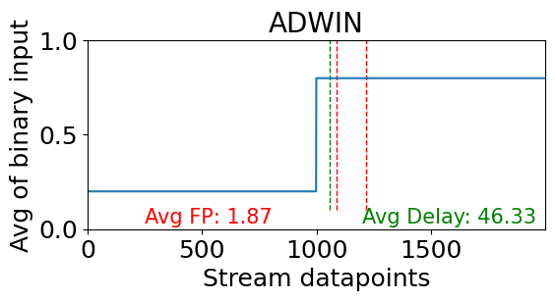}} \\
    \subfigure{\includegraphics[width=1\columnwidth]{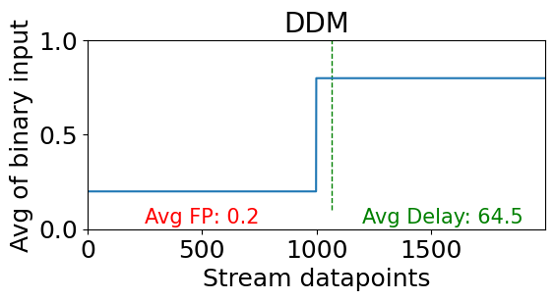}} \\
    \subfigure{\includegraphics[width=1\columnwidth]{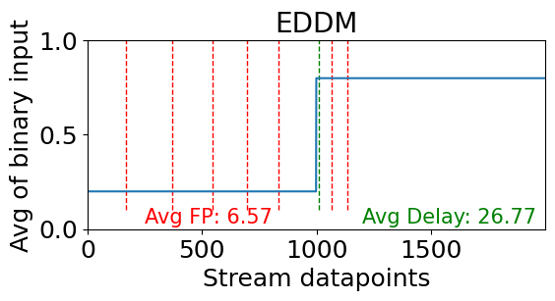}} \\
    \subfigure{\includegraphics[width=1\columnwidth]{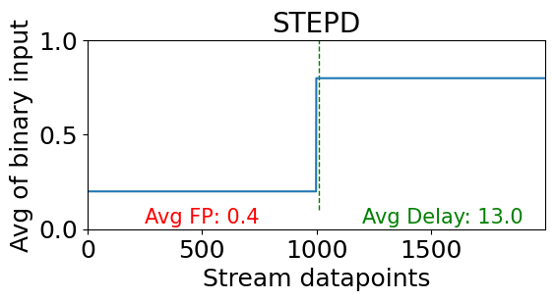}} \\
    \subfigure{\includegraphics[width=1\columnwidth]{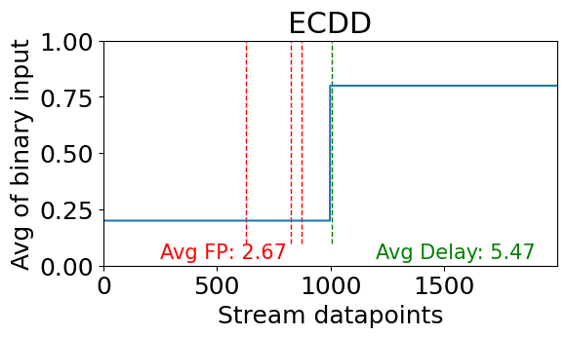}} \\
\end{multicols}
\caption{Sudden binary drift detection with average FP rates compared to drift-detection delays.}
\label{fig:abrupt}
\end{figure}
}

{
\setlength{\belowcaptionskip}{-15pt}
\setlength{\abovecaptionskip}{-10pt}
\begin{figure}
\setlength{\columnsep}{0.1\columnsep}
\setlength{\subfigtopskip}{0.1\subfigtopskip}
\setlength{\subfigbottomskip}{0.1\subfigbottomskip}
\setlength{\subfigcapskip}{0.1\subfigcapskip}
\begin{multicols}{2}
    \subfigure{\includegraphics[width=1\columnwidth]{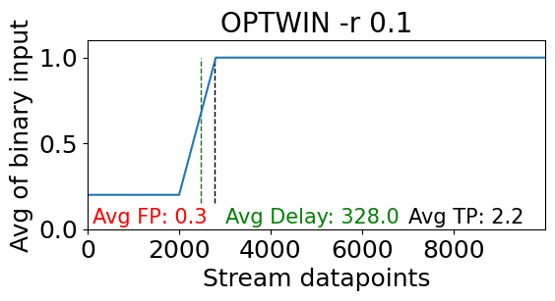}} \\
    \subfigure{\includegraphics[width=1\columnwidth]{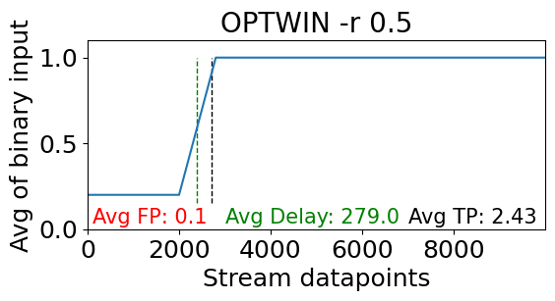}} \\
    \subfigure{\includegraphics[width=1\columnwidth]{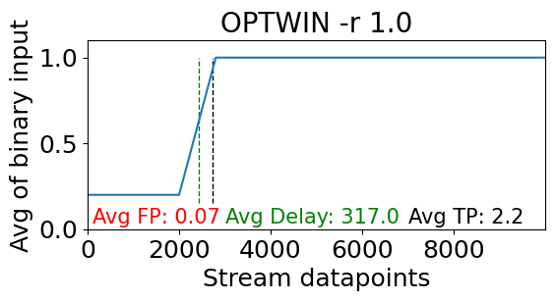}} \\
    \subfigure{\includegraphics[width=1\columnwidth]{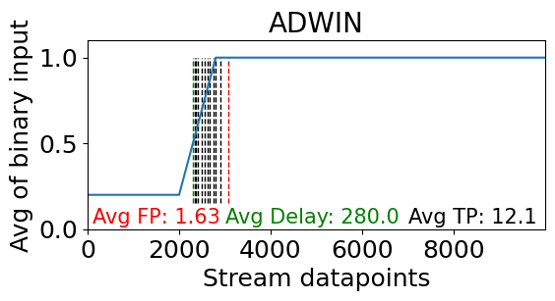}} \\
    \subfigure{\includegraphics[width=1\columnwidth]{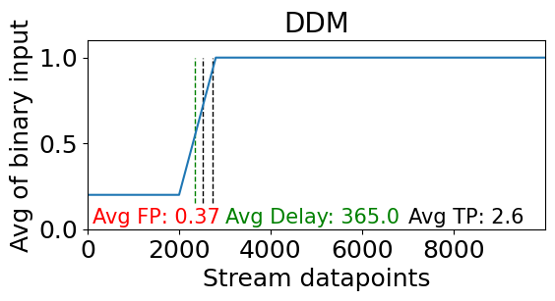}} \\
    \subfigure{\includegraphics[width=1\columnwidth]{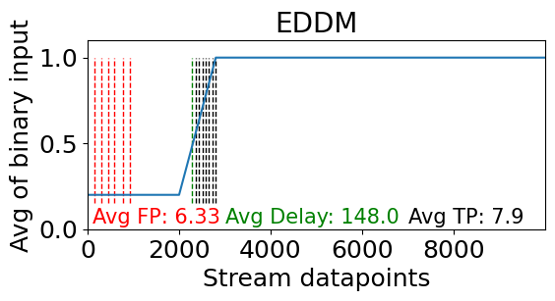}} \\
    \subfigure{\includegraphics[width=1\columnwidth]{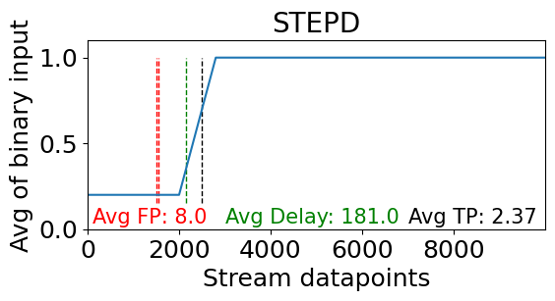}} \\
    \subfigure{\includegraphics[width=1\columnwidth]{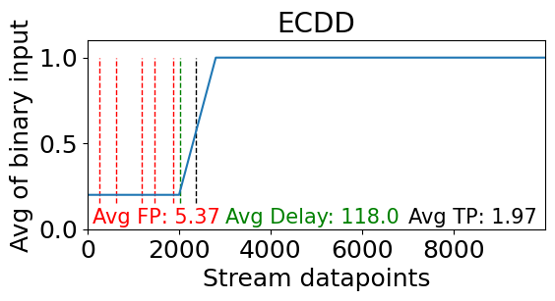}} \\
\end{multicols}
\caption{Gradual binary drift detection with average TP and FP rates compared to drift-detection delays.}
\label{fig:gradual}
\end{figure}
}

In Figure \ref{fig:agrawl}, we can once again see OPTWIN and DDM identifying all the drifts with a low FP rate. However, in this scenario, ADWIN also produces only a few FPs, which we noticed to occur immediately after correctly identifying a concept drift. This is due to the time it takes to adapt its window size and remove data points from before the concept drift. In this use case, ECDD and STEPD produce almost ``random guesses'' on the location of the concept drifts. 

{
\setlength{\abovecaptionskip}{-10pt}
\setlength{\belowcaptionskip}{-18pt}
\begin{figure}
\setlength{\columnsep}{0.1\columnsep}
\setlength{\subfigtopskip}{0.1\subfigtopskip}
\setlength{\subfigbottomskip}{0.1\subfigbottomskip}
\setlength{\subfigcapskip}{0.1\subfigcapskip}
\begin{multicols}{2}
    \subfigure{\includegraphics[width=1\columnwidth]{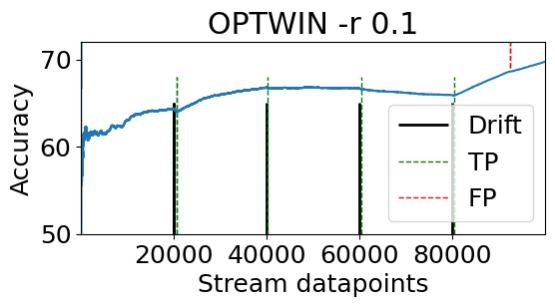}} \\
    \subfigure{\includegraphics[width=1\columnwidth]{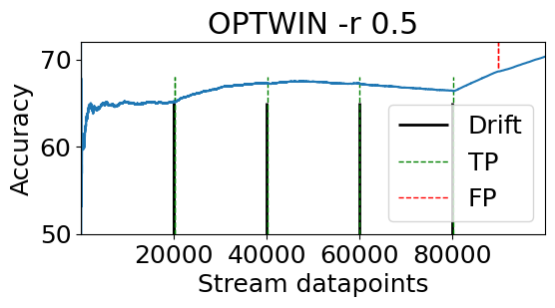}} \\
    \subfigure{\includegraphics[width=1\columnwidth]{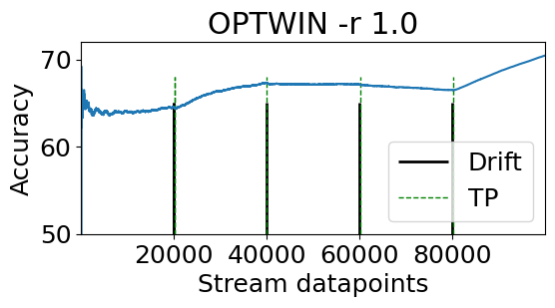}} \\
    \subfigure{\includegraphics[width=1\columnwidth]{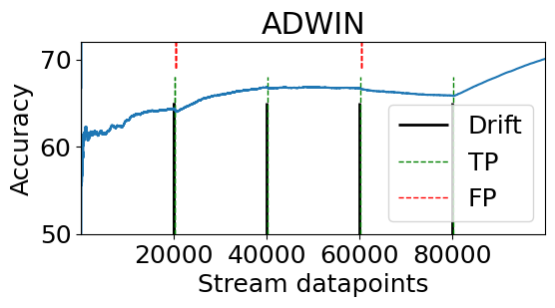}} \\
    \subfigure{\includegraphics[width=1\columnwidth]{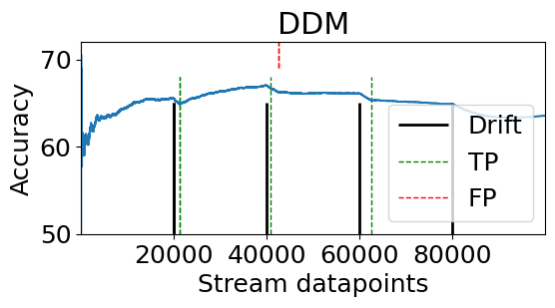}} \\
    \subfigure{\includegraphics[width=1\columnwidth]{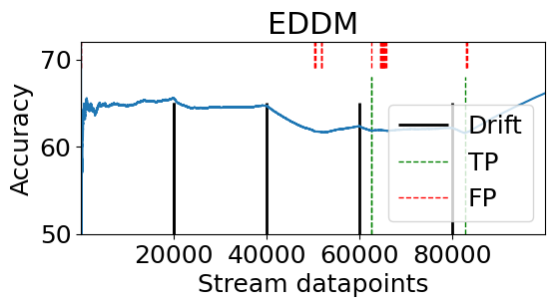}} \\
    \subfigure{\includegraphics[width=1\columnwidth]{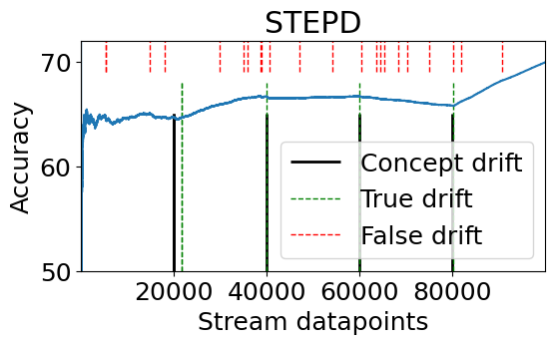}} \\
    \subfigure{\includegraphics[width=1\columnwidth]{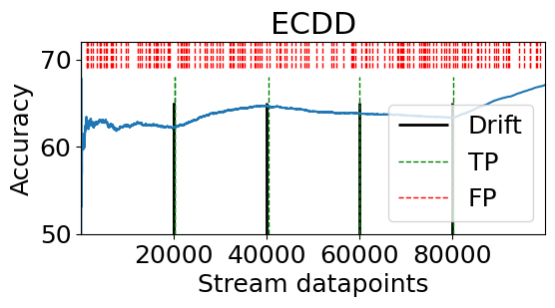}} \\
\end{multicols}
\caption{TP and FP rates for drift detection on the AGRAWAL dataset with sudden concept drifts.}
\label{fig:agrawl}
\end{figure}
}

\subsection{Classification Experiments}

In the ``Classification'' experiments (still using MOA as platform), we can analyze the average accuracy achieved by the NB classifier when varying the drift detectors (cf. Table \ref{tb:adaptation_sudden}). The idea is that the better a drift detector's performance, the better the classifier can adapt to the concept drifts, thus generating a higher prediction accuracy. However, the fast and accurate detection of the drifts did not always traverse to a better accuracy for the classifier. We can observe this by comparing Tables \ref{tb:false_positive} and \ref{tb:adaptation_sudden}, in which experiments using drift detectors that produced a low F1-score in Table \ref{tb:false_positive} achieved good accuracy in Table \ref{tb:adaptation_sudden}. Moreover, the drift detectors with the best accuracy on real-world data sets were the ones that detected more drifts, which is why ECDD achieved such good accuracy (being the detector that produced the higher amount of FPs). For example, ECDD detected 426 drifts on the Electricity dataset---more than twice as many as all other drift detectors. 

{
\small
\setlength{\abovecaptionskip}{-15pt}
\begin{table*}
\centering
\begin{tabular}{ l | c c c | c c c | c c }
& \multicolumn{3}{c}{Sudden drift} & \multicolumn{3}{c}{Gradual drift} & \multicolumn{2}{c}{Real-world datasets} \\
Drift Detector & STAGGER & Random RBF & AGRAWAL & STAGGER & Random RBF & AGRAWAL & Electricity & Covertype \\
\hline
No drift detector & 66.31 & 58.69 & 57.94 & 66.31 & 58.69 & 57.93 & 73.36 & 60.52 \\
ADWIN & 99.89 & 69.74 & \textbf{70.22} & 98,78 & 69,48 & 69,89  & 80.01 & 82.49\\
DDM & \textbf{99.96} & 68.18 & 65.53 & 98,67 & 68,19 & 65,80 & 81.18 & 88.03\\
EDDM & 99.87 & 65.99 & 65.46 & \textbf{98,83} & 66,92 & 65,31 & 84.83 & 86.08\\
STEPD & 98.74 & \textbf{70.92} & 69.48 & 98,78 & \textbf{70,49} & 69,13 & 84.49 & 87.53\\
ECDD & \textbf{99.96} & 68.64 & 67.03 & 98,71 & 68,52 & 66,80 & \textbf{86.76} & \textbf{90.16}\\
OPTWIN $\rho=0.1$ & \textbf{99.96} & 69.65 & 70.11 & 98,72 & 69,54 &  \textbf{69,94} & 79.99 & 83.29\\
OPTWIN $\rho=0.5$ & \textbf{99.96} & 69.77 & 69.91 & 98,51 & 69,48 &  69,27 & 83.30 & 85.63\\
OPTWIN $\rho=1.0$ & \textbf{99.96} & 68.67 & 69.44 & 98,36 & 68,48 &  68,61 & 82.96 & 86.31\\
\end{tabular}
\caption{Accuracy of NB on synthetic and real-world datasets.}
\label{tb:adaptation_sudden}
\end{table*}
}

\subsection{Neural Network Experiments}

To further explore OPTWIN's behavior in a regression scenario, we compared it with ADWIN for identifying drifts from the loss of a CNN. We chose ADWIN as our baseline because it was the drift detector with the best F1-score among the ones that do not require binary inputs (thus excluding DDM, EDDM and ECDD). To simplify the reader's understanding of the generation of the concept drift, instead of training a regression problem on a CNN, we here focused on image classification by swapping the labels of the images. Thus, we provoked 4 concept drifts by swapping the labels of two classes of images every 20\% of the simulated data stream. For example, after 62,480 iterations, we swapped the labels between images from ``cats'' to ``horses''. Nevertheless, as the variable tracked by the drift detectors is the loss of the CNN, the type of the problem or the network architecture should not affect the experiment.

We pre-trained an image classification model \cite{tensorflow_2022} using the CIFAR-10 \cite{krizhevsky2009learning} data set during 100 epochs, achieving an average of 89\% training accuracy over 3 different runs. Then, we simulated an OL scenario with concept drifts. Our data stream was formed by batches of 32 images from the CIFAR-10 dataset; with a total of 312,400 data points (equivalent to 100 epochs). We simulated our 4 concept drifts by swapping the labels of two classes every 62,480 iterations (the equivalent of 20 epochs). At every iteration, the model classified the 32 images and outputted the loss of the batch. We inputted this loss into the drift-detection algorithm. If a drift was detected, the next 9,372 batches of images (the equivalent of 3 epochs) were used for fine-tuning the model (thus adapting it to the concept drift). Therefore, the goal was for the drift detector to identify the 4 concept drifts that we simulated, triggering the fine-tuning of the model for a total of 12 epochs (3 epochs per drift).

In Figure \ref{fig:cifar}, we compare OPTWIN and ADWIN under the setting described above. First, we note that ADWIN's high FP rates made it retrain the model for much longer than OPTWIN. In comparison, ADWIN identified 15 concept drifts (with 11 FPs), thus triggering model fine-tuning for 61,562 iterations which took in total 945 seconds. In contrast, OPTWIN identified just 5 drifts (with 1 FP), thus triggering the model fine-tuning for 23,430 iterations which took 781 seconds. We note that OPTWIN's running time per iteration is superior to ADWIN, $1e^{-5}$ against $6e^{-6}$ seconds. However, OPTWIN still ends up speeding up the OL pipeline whenever retraining is triggered by the concept-drift detection (21\% faster in this use case). This is because the training of a learner is usually computationally more expensive than the drift detection. Thus, with fewer FPs, the total training time can be reduced substantially.

{
\setlength{\belowcaptionskip}{-15pt}
\setlength{\abovecaptionskip}{-10pt}
\begin{figure}
\setlength{\columnsep}{0.1\columnsep}
\setlength{\subfigtopskip}{0.1\subfigtopskip}
\setlength{\subfigbottomskip}{0.1\subfigbottomskip}
\setlength{\subfigcapskip}{0.1\subfigcapskip}
\begin{multicols}{2}
    \subfigure{\includegraphics[width=.98\columnwidth]{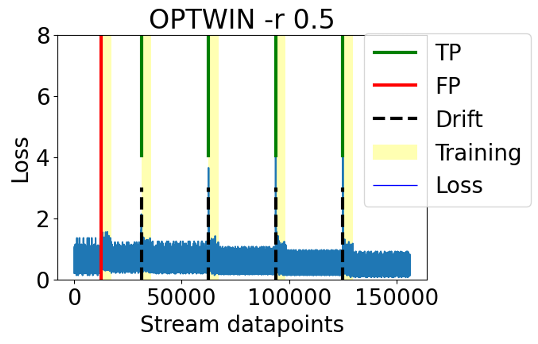}} \\
    \subfigure{\includegraphics[width=.98\columnwidth]{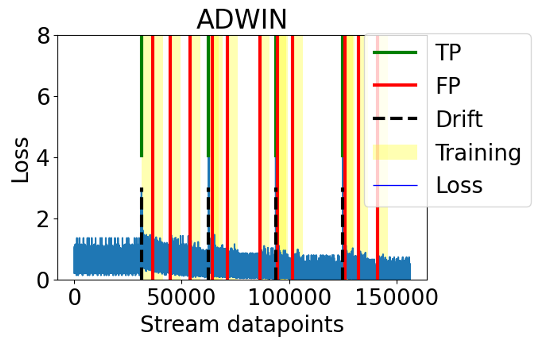}} \\
\end{multicols}
\caption{Sudden drift detection over the loss of a NN.}
\label{fig:cifar}
\end{figure}
}
\section{Conclusion}

In this paper, we presented OPTWIN, a novel concept-drift detector that uses a sliding window of errors to identify concept drifts with a low FP rate. OPTWIN's novelty relies on the assumption that changes also in the variances of the elements ingested from a data stream can be an indication of a concept drift. This assumption lets it optimally divide its sliding window and apply the $t$- and $f$-tests to determine whether a concept drift occurred. To assess OPTWIN, we compared it with 5 popular drift detectors in 11 different experiments. As a result, OPTWIN achieved higher F1-scores in most of our experiments. In fact, OPTWIN had the best F1-score (with statistical significance) while maintaining a similar drift-detection delay compared to the drift detectors suited for both classification and regression problems. Moreover, OPTWIN could speed up the overall OL pipeline of a CNN by 21 (compared to ADWIN) due to its low FP rate. To conclude, we conjecture that OPTWIN's characteristics can enable even higher speed-ups in scenarios where the drift identification triggers the re-training of complex models.

\begin{acks}
This work is funded by the Luxembourg National Research Fund under the PRIDE program (PRIDE17/12252781).
\end{acks}

\bibliographystyle{ACM-Reference-Format}
\bibliography{bibliography}

\end{document}